\documentclass[letterpaper, 10 pt, conference]{ieeeconf}

\IEEEoverridecommandlockouts
\pdfoutput=1

\usepackage{xparse}
\usepackage{amsmath}
\usepackage{amssymb}
\usepackage{amsfonts}
\usepackage{bbm}
\usepackage{color}

\usepackage{graphicx}
\usepackage[keeplastbox]{flushend}

\usepackage{hyperref}
\usepackage{cleveref}
\usepackage{subcaption}

\usepackage{url}
\usepackage{framed}

\usepackage{algpseudocode}
\usepackage{algorithm}

\usepackage{threeparttable}
\usepackage{booktabs}
\usepackage{arydshln}

\newcommand\T{\rule{0pt}{2.6ex}}        
\newcommand\B{\rule[-1.2ex]{0pt}{0pt}}  

\NewDocumentCommand\SunDirectionGT{}{\mbf s_k}
\NewDocumentCommand\SunDirectionGTAll{}{\mbf S}
\NewDocumentCommand\SunImages{}{\mbf X}

\NewDocumentCommand\SunDirectionEst{}{\mbfhat s_k}
\NewDocumentCommand\SunDirectionEstNorm{}{\mbfhat s_k}
\NewDocumentCommand\SunDirectionEstMean{}{\bar{\mbfhat s}_k}
\NewDocumentCommand\CNNLoss{m}{\mathcal{L}(#1)}

\NewDocumentCommand\Transform{}{\mbf{T}}

\newcommand{\mbf}[1]{\mathbf{#1}}

\newcommand{\mbfhat}[1]{\hat{\mbf{#1}}}

\newcommand{\mbs}[1]{\boldsymbol{#1}}

\newcommand{\One}{\mbf{1}}

\DeclareMathAlphabet{\mbfh}{OML}{cmm}{b}{it}

\newcommand{\cframe}[1]{\ensuremath \underrightarrow{\mathcal{F}}_{#1}}

\newcommand{\norm}[1]{\left\Vert#1\right\Vert}

\newcommand{\bbm}{\begin{bmatrix}}
\newcommand{\ebm}{\end{bmatrix}}

\newcommand{\set}[1]{\left\{#1\right\}}

\newcommand{\SE}[1]{SE(#1)}

\newcommand{\diag}[1]{\textup{diag} \left\{ #1 \right\}}

\title{\LARGE \bf Reducing Drift in Visual Odometry by Inferring Sun Direction\\Using a Bayesian Convolutional Neural Network$^\ddagger$}

\author{Valentin Peretroukhin$^\dagger$, Lee Clement$^\dagger$, and Jonathan Kelly\thanks{$^\dagger$Valentin Peretroukhin and Lee Clement contributed equally to this work and jointly assert first authorship. All authors are with the Space \& Terrestrial Autonomous Robotic Systems (STARS) laboratory at the University of Toronto Institute for Aerospace Studies (UTIAS), Canada {\tt \{lee.clement, v.peretroukhin\}@mail.utoronto.ca, jkelly@utias.utoronto.ca}.\newline $^\ddagger$This version of the paper corrects minor errors in \Cref{sec:sun-bcnn-test-results} and the caption of \Cref{fig:error_over_time} of the published version.}}

\begin{document}

\maketitle
\thispagestyle{empty}
\pagestyle{empty}

\begin{abstract}
We present a method to incorporate global orientation information from the sun into a visual odometry pipeline using only the existing image stream, where the sun is typically not visible. We leverage recent advances in Bayesian Convolutional Neural Networks to train and implement a sun detection model that infers a three-dimensional sun direction vector from a single RGB image. Crucially, our method also computes a principled uncertainty associated with each prediction, using a Monte Carlo dropout scheme. We incorporate this uncertainty into a sliding window stereo visual odometry pipeline where accurate uncertainty estimates are critical for optimal data fusion. Our Bayesian sun detection model achieves a median error of approximately 12 degrees on the KITTI odometry benchmark training set, and yields improvements of up to 42\% in translational ARMSE and 32\% in rotational ARMSE compared to standard VO. An open source implementation of our Bayesian CNN sun estimator (Sun-BCNN) using Caffe is available at \normalfont{\url{https://github.com/utiasSTARS/sun-bcnn-vo}}.
 \end{abstract}

\section{Introduction}
 Egomotion estimation is a fundamental building block of mobile autonomy. Although there exist an array of possible algorithm-sensor combinations that can estimate motion in unknown environments (e.g., LIDAR-based point-cloud matching \cite{Zhang2015-gf} and visual-inertial navigation \cite{Leutenegger2015-fk}), egomotion estimation remains a dead-reckoning technique that accumulates unbounded estimation error over time in the absence of global information such as GPS or a known map.

In this work, we focus on one technique to infer global orientation information without a known map: computing the direction of the sun. By leveraging recent advances in Bayesian Convolutional Neural Networks (BCNNs), we demonstrate how we can train a deep model to compute a direction vector from a single RGB image using only 20,000 training images. Furthermore, we show that our network can produce a principled covariance estimate that can readily be used in an egomotion estimation pipeline. We demonstrate one such use by incorporating sun direction estimates into a stereo visual odometry (VO) pipeline and report significant error reductions of up to 42\% in translational average root mean squared error (ARMSE) and 32\% in rotational ARMSE compared to plain VO on the KITTI odometry benchmark training set \cite{Geiger2013-ky}.
\vspace{1cm}

Our main contributions are as follows:
\begin{enumerate}
\item We apply a Bayesian CNN to the problem of sun direction estimation, incorporating the resulting covariance estimates into a visual odometry pipeline;
\item We show that a Bayesian CNN with dropout layers after each convolutional and fully-connected layer can achieve state-of-the-art accuracy at test time;
\item We learn a 3D unit-length sun direction vector, appropriate for full 6-DOF pose estimation;
\item We present experimental results on 21.6 km of urban driving data from the KITTI odometry benchmark training set~\cite{Geiger2013-ky}; and
\item We release our Bayesian CNN sun estimator (Sun-BCNN) as open-source code.
\end{enumerate}

\section{Related Work}
Visual odometry (VO), a technique to estimate the egomotion of a moving platform equipped with one or more cameras, has a rich history of research including a notable implementation onboard the Mars Exploration Rovers (MERs) \cite{Scaramuzza2011-qr}. Modern approaches to VO can achieve estimation errors below 1\% of total distance traveled \cite{Geiger2013-ky}. To achieve such accurate  and robust estimates, modern techniques use careful visual feature pruning \cite{Cvisic2015-mt}, adaptive robust methods \cite{Alcantarilla2016-fs,Peretroukhin2016-om}, or operate directly on pixel intensities \cite{Engel2015-il}.

\begin{figure}[t]
    \centering
      \includegraphics[width=0.5\textwidth]{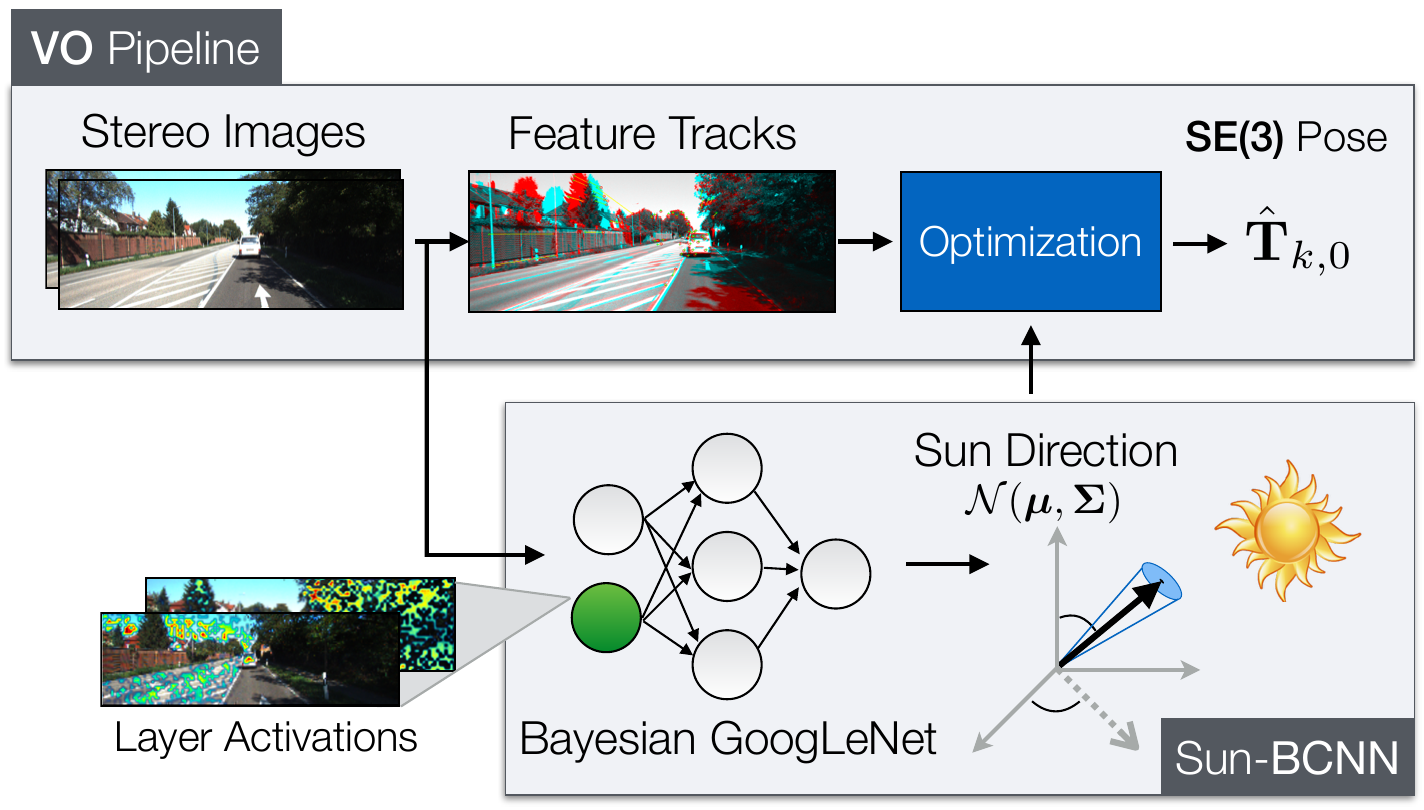}
      \caption{Sun-BCNN (Sun Bayesian Convolutional Neural Network) incorporated into a visual odometry (VO) pipeline. A Bayesian CNN infers sun direction estimates as a mean and covariance, which are then incorporated into a sliding window bundle adjuster to produce a final trajectory estimate.}
    \label{fig:system}
\end{figure}

\begin{figure*}
    \centering
    \includegraphics[width=0.98\textwidth]{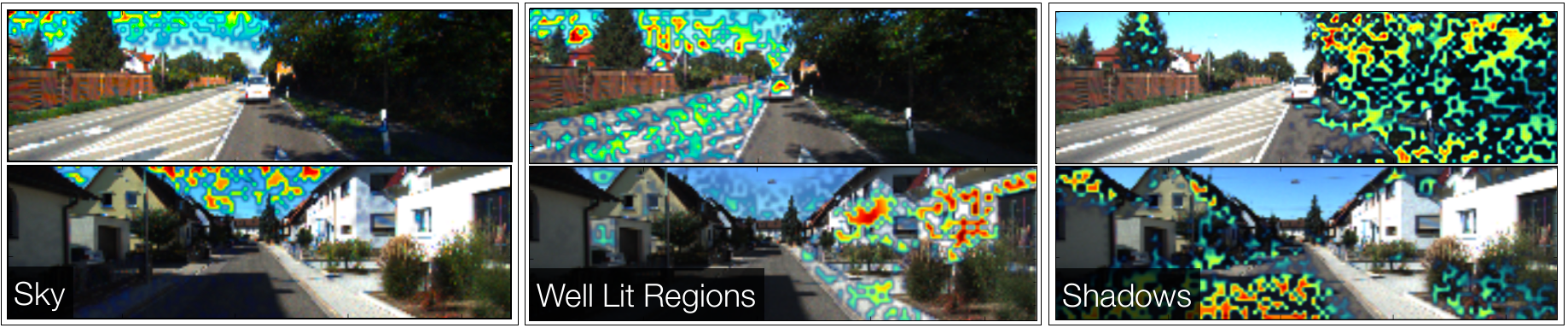}
    \caption{Three \texttt{conv1} layer activation maps superimposed on two images from the KITTI \cite{Geiger2013-ky} odometry benchmark  \texttt{00} and \texttt{04} for three selected filters. Each filter picks out salient parts of the image that aid in sun direction inference.}
    \label{fig:cnn_activations}
\end{figure*}

Independent of the estimator, VO exhibits super-linear error growth  \cite{Olson2003-ax}, and is particularly sensitive to errors in orientation \cite{Olson2003-ax, Cvisic2015-mt}. One way to reduce orientation error is to incorporate observations of a landmark whose position or direction in the navigation frame is known \emph{a priori}. The sun is an example of such a known directional landmark. Accordingly, sun sensors have been used to improve the accuracy of VO in planetary analogue environments \cite{Furgale2011-zu,Lambert2012-um}, and were also incorporated into the MERs \cite{Maimone2007-tc,Eisenman2002-cg}. More recently, software-based alternatives have been developed that can estimate the direction of the sun from a single image, making sun-aided navigation possible without additional sensors or a specially-oriented camera~\cite{Clement2016-ir}. Some of these methods have been based on hand-crafted illumination cues~\cite{Lalonde2011-jw,Clement2016-ir}, while others have attempted to learn such cues from data using deep Convolutional Neural Networks (CNNs)~\cite{Ma2016-at}.

CNNs have been applied to a wide range of classification, segmentation, and learning tasks in computer vision~\cite{LeCun2015-qf}. Recent work has shown that CNNs can learn orientation information directly from images by modifying the loss functions of existing discrete classification-based CNN architectures into continuous regression losses~\cite{Ma2016-at, kendall2015posenet, Kendall2016-zf}. Despite their success in improving prediction accuracy, most existing CNN-based models do not report principled uncertainty estimates, which are important in the context of data fusion.  To address this, Gal and Ghahramani~\cite{Gal2016-ny} showed that it is possible to achieve principled covariance outputs with only minor modifications to existing CNN architectures. An early application of this uncertainty quantification was presented by Kendall et al. \cite{Kendall2016-zf} who used it to improve their prior work on camera pose regression.

Our method is similar in spirit to the work of Ma et al. \cite{Ma2016-at} who built a CNN-based sun sensor as part of a relocalization pipeline. We also extend the work of Clement et al. \cite{Clement2016-ir} who demonstrated that virtual sun sensors can improve VO accuracy. Our model makes three important improvements: 1) in addition to a point estimate of the sun direction, we output a principled covariance estimate that is incorporated into our estimator; 2) we produce a full 3D sun direction estimate with azimuth and zenith angles that is better suited to 6-DOF estimation problems (as opposed to only the azimuth angle and 3-DOF estimator in \cite{Ma2016-at}); and 3) we incorporate the sun direction covariance into a VO estimator that accounts for growth in pose uncertainty over time (unlike \cite{Clement2016-ir}). Furthermore, our Bayesian CNN includes a dropout layer after every convolutional and fully connected layer (as outlined in \cite{Gal2016-ny} but not done in \cite{Kendall2016-zf}), which produces more principled covariance outputs.

\section{Indirect Sun Detection using a Bayesian Convolutional Neural Network} \label{sec:sun-bcnn}

We use a Convolutional Neural Network (CNN) to infer the direction of the sun. We motivate the choice of a deep model through the empirical findings of Clement et al. \cite{Clement2016-ir} and Ma et al.~\cite{Ma2016-at}, who demonstrated that a CNN-based sun detector can substantially outperform hand-crafted models such as that of Lalonde et al.~\cite{Lalonde2011-jw}.

We choose a deep neural network structure based on GoogLeNet~\cite{Szegedy2015-uw} due to its use in past work that adapted it for orientation regression~\cite{kendall2015posenet}. Unlike Ma et al. \cite{Ma2016-at}, we choose to transfer weights trained on the MIT Places dataset~\cite{zhou2014MITPlaces} rather than ImageNet~\cite{deng2009imagenet}.
We believe the MIT Places dataset is a more appropriate starting point for localization tasks than ImageNet since it includes outdoor scenes and is concerned with classifying physical locations rather than objects.

\subsection{Cost Function}
We train the network by minimizing the cosine distance between the (unit-norm) target sun direction vector $\SunDirectionGT$ and the predicted (unit-norm) sun direction vector $\SunDirectionEst$, where $k$ indexes the images in the training set:
\begin{equation}
	\CNNLoss{\SunDirectionEst} = 1 - (\SunDirectionEst \cdot \SunDirectionGT) ,
	\label{eq:cnn_loss}
\end{equation}

Note that in our implementation, we do not formulate the cosine distance loss explicitly, but instead minimize half the square of the Euclidian distance between $\SunDirectionGT$ and $\SunDirectionEst$. Since both vectors have unit length, this is equivalent to minimizing \Cref{eq:cnn_loss}:
 \begin{align*}
 	\frac{1}{2} \norm{\SunDirectionEstNorm  - \SunDirectionGT}^2 &= \frac{1}{2} \left( \norm{\SunDirectionEstNorm}^2 + \norm{\SunDirectionGT}^2 - 2 (\SunDirectionEstNorm \cdot \SunDirectionGT) \right) \\
 		&= 1 - (\SunDirectionEstNorm \cdot \SunDirectionGT) \\
 		&= \CNNLoss{\SunDirectionEst}.
 \end{align*}

\subsection{Uncertainty Estimation}
 To output principled covariances for sun direction estimates, we adopt Bayesian Convolution Neural Networks (BCNNs)~\cite{Gal2016-ny, Gal2016CNN, Gal2016UncertaintyThesis}. BCNNs rely on a connection between stochastic regularization (e.g. dropout, a widely adopted technique in deep learning) and approximate variational inference of a Bayesian Neural Network. We outline the technique here briefly, and refer the reader to~\cite{Gal2016CNN} for more details.

The method begins with a prior on the weights in a deep neural network, $p(\mbs w)$, and attempts to compute a posterior distribution $p(\mbs w | \mbf \SunImages, \mbf \SunDirectionGTAll)$ given training inputs $\SunImages$ and targets $\SunDirectionGTAll = \set{\SunDirectionGT}$. This posterior can be used to compute a predictive distribution for test samples but is generally intractable. To overcome this, the BCNN approach notes that CNN training with stochastic regularization can be viewed as variational inference if we define a variational distribution $q(\mbs w)$ as:
\begin{align}
	q(\mbs w_i) &= \mbf{M}_i ~ \diag{\{b^i_j\}_{j=1}^{K_i}}, \\
    b^i_j &\in \text{Bernoulli}(p_i).
\end{align}
Here, $i$ indexes a particular layer in the neural network with $K_i$ weights, $\mbf M_i$ are the weights to be optimized, $b^i_j$ are Bernoulli distributed binary variables, and $p_i$ is the dropout probability for weights in layer $i$.

With this variational distribution $q(\mbs w)$, training a CNN with dropout results in the same $\mbs w$ as minimizing the Kullback-Leibler (KL) divergence between the variational distribution and the true posterior: $\text{KL}(p(\mbs w | \SunImages, \mbf \SunDirectionGTAll) || q(\mbs w))$.
At test time, the first two moments of the predictive distribution are approximated using Monte Carlo integration over the weights $\mbs w$:
\begin{align}
\label{eq:sun_direction_mean}
\mathbb{E} (\SunDirectionEst^*) &= \SunDirectionEstMean^* \approx \frac{1}{N} \sum_{n=1}^N \SunDirectionEst^* (\mbf x^*, \mbs{w}^n), \\
 \mathbb{E} (\SunDirectionEst^* \SunDirectionEst^{*T}) &\approx \tau^{-1}\mbf 1
 +  \frac{1}{N} \sum_{n=1}^N \SunDirectionEst^*(\mbf x^*, \mbs w^n)\SunDirectionEst^*(\mbf x^*, \mbs w^n)^T \notag \\ &-  \SunDirectionEstMean^*\SunDirectionEstMean^{*T},
 \label{eq:bcnn_covar}
  \end{align} where $\One$ is the identity matrix, and $\mbs w^n$ is a sample from $q(\mbs w)$ (obtained by sampling the network with dropout). The model precision, $\tau$, is computed as
\begin{equation}
	\label{eq:model_precision}
	\tau = \frac{p l^2}{2 M \lambda},
\end{equation}
where $p$ is the dropout probability, $l$ is the characteristic length scale, $M$ is the number of samples in the training data, and $\lambda$ is the weight decay.

Following Gal and Ghahramani~\cite{Gal2016CNN}, we build our BCNN by adding dropout layers after every convolutional and fully connected layer in the network. We then retain these layers at test time to sample the network stochastically (following the technique of Monte Carlo Dropout), and obtain the relevant statistical quantities using \Cref{eq:bcnn_covar,eq:sun_direction_mean}.

\section{Sliding Window Stereo Visual Odometry} \label{sec:stereo-vo}
We adopt a sliding window sparse stereo VO technique that has been used in a number of successful mobile robotics applications~\cite{Cheng2006-nl,Furgale2010-to,Geiger2011-xe,Kelly2008-mh}.
Our task is to estimate a window of $\SE{3}$ poses $\set{\Transform_{k_1,0}, \dots, \Transform_{k_2,0}}$ expressed in a base coordinate frame $\cframe{0}$, given a prior estimate of the transformation $\Transform_{k_1,0}$.
We accomplish this by tracking keypoints across pairs of stereo images and computing an initial guess for each pose in the window using frame-to-frame point cloud alignment, which we then refine by solving a local bundle adjustment problem over the window.
In our experiments we choose a window size of two, which provides good VO accuracy at low computational cost.
As discussed in \Cref{sec:orientation}, we select the initial pose $\Transform_{1,0}$ to be the first GPS ground truth pose such that $\cframe{0}$ is a local East-North-Up (ENU) coordinate system with its origin at the first GPS position.

\subsection{Observation Model}
We assume that our stereo images have been de-warped and rectified in a pre-processing step, and model the stereo camera as a pair of perfect pinhole cameras with focal lengths $f_u, f_v$ and principal points $\left(c_u,c_v\right)$, separated by a fixed and known baseline $b$.
If we take $\mbf{p}_0^j$ to be the homogeneous 3D coordinates of keypoint $j$, expressed in our chosen base frame $\cframe{0}$, we can transform the keypoint into the camera frame at pose $k$ to obtain $\mbf{p}_k^j = \Transform_{k,0}\mbf{p}_0^j = \bbm p_{k,x}^j & p_{k,y}^j & p_{k,z}^j & 1 \ebm^T$. Our observation model $\mbf{g}\left(\cdot\right)$ can then be formulated as
\begin{align} \label{eq:cam_model}
    \mbf{y}_{k,j} &= \mbf{g}\left(\mbf{p}_k^j\right)
                    = \bbm u \\ v \T \\ d \T \ebm
                    = \bbm
			   		    f_u p_{k,x}^j / p_{k,z}^j + c_u \\
			   		    f_v p_{k,y}^j / p_{k,z}^j + c_v \T \\
			   		    f_u b / p_{k,z}^j  \T
			         \ebm,
\end{align}
where $\left(u,v\right)$ are the keypoint coordinates in the left image and $d$ is the disparity in pixels.

\subsection{Sliding Window Bundle Adjustment}
We use the open-source \texttt{libviso2} package~\cite{Geiger2011-xe} to detect and track keypoints between stereo image pairs.
Based on these keypoint tracks, a three-point Random Sample Consensus (RANSAC) algorithm generates an initial guess of the inter-frame motion and rejects outlier keypoint tracks by thresholding their reprojection error.
We compound these pose-to-pose transformation estimates through our chosen window and refine them using a local bundle adjustment, which we solve using the nonlinear least-squares solver Ceres~\cite{ceres-solver}.
The objective function to be minimized can be written as
\begin{equation} \label{eq:cost_function}
    \mathcal{J} = \mathcal{J}_{\text{reprojection}} + \mathcal{J}_{\text{prior}},
\end{equation}
where
\begin{equation} \label{eq:reprojection_cost}
	\mathcal{J}_{\text{reprojection}} = \sum_{k=k_1}^{k_2} \sum_{j=1}^J \mbf{e}_{\mbf{y}_{k,j}}^T \mbf{R}^{-1}_{\mbf{y}_{k,j}} \mbf{e}_{\mbf{y}_{k,j}}
\end{equation}
and
\begin{equation} \label{eq:prior_cost}
	\mathcal{J}_{\text{prior}} = \mbf{e}_{\check{\Transform}_{k_1,0}}^T \mbf{R}^{-1}_{\check{\Transform}_{k_1,0}} \mbf{e}_{\check{\Transform}_{k_1,0}}.
\end{equation}

The quantity $\mbf{e}_{\mbf{y}_{k,j}} = \mbfhat{y}_{k,j} - \mbf{y}_{k,j}$ represents the reprojection error of keypoint $j$ for camera pose $k$, with $\mbf{R}_{\mbf{y}_{k,j}}$ being the covariance of these errors.
The predicted measurements are given by $\mbfhat{y}_{k,j} = \mbf{g}\left(\mbfhat{T}_{k,0} \mbfhat{p}^j_{0}\right)$, where $\mbfhat{T}_{k,0}$ and $\mbfhat{p}^j_{0}$ are the estimated poses and keypoint positions in base frame $\cframe{0}$.

The cost term $\mathcal{J}_{\text{prior}}$ imposes a normally distributed prior $\check{\Transform}_{k_1,0}$ on the first pose in the current window, based on the estimate of this pose in the previous window.
The error in the current estimate $\hat{\Transform}_{k_1,0}$ of this pose compared to the prior can be computed using the $\SE{3}$ matrix logarithm as $\mbf{e}_{\check{\Transform}_{k_1,0}} = \log \left( \check{\Transform}_{k_1,0}^{-1}\hat{\Transform}_{k_1,0} \right)$.
The $6 \times 6$ matrix $\mbf{R}_{\check{\Transform}_{k_1,0}}$ is the covariance associated with $\check{\Transform}_{k_1,0}$ in its local tangent space, and is obtained as part of the previous window's bundle adjustment solution.
This prior term allows consecutive windows of pose estimates to be combined in a principled way that appropriately propagates global pose uncertainty from window to window, which is essential in the context of optimal data fusion.

\subsection{Sun-based Orientation Correction} \label{sec:orientation}
In order to combat drift in the VO estimate produced by accumulated orientation error, we adopt the technique of Lambert et al.~\cite{Lambert2012-um} to incorporate absolute orientation information from the sun directly into the estimation problem.
We assume the initial camera pose and its timestamp are available from GPS and use them to determine the global direction of the sun $\mbf{s}_0$, expressed as a 3D unit vector, based on a solar ephemeris model that computes the sun direction for a given date, time, and location on Earth.
We define the world frame $\cframe{0}$ to be a local ENU coordinate system with the initial GPS position as its origin.
At each timestep we update $\mbf{s}_0$ by querying the ephemeris model using the current timestamp and the initial camera pose, allowing us to account for the apparent motion of the sun over long trajectories.
Note that here we are using the notation $\mbf{s}_k$ to represent the sun vector predicted by our sun sensing apparatus (denoted $\SunDirectionEst$ in \Cref{sec:sun-bcnn}), not the ground truth training vector.

By transforming the global sun direction into each camera frame $\cframe{k}$ in the window, we obtain expected sun directions $\mbfhat{s}_k = \mbfhat{T}_{k,0} \mbf{s}_0$, where $\mbfhat{T}_{k,0}$ is the current estimate of camera pose $k$ in the base frame.
We compare the expected sun direction $\mbfhat{s}_k$ to the estimated sun direction $\mbf{s}_k$ to introduce an additional error term into the bundle adjustment cost function (cf. \Cref{eq:cost_function}):
\begin{equation} \label{eq:cost_function_with_sun}
    \mathcal{J} = \mathcal{J}_{\text{reprojection}} + \mathcal{J}_{\text{prior}} + \mathcal{J}_{\text{sun}},
\end{equation}
where
\begin{equation} \label{eq:sun_cost}
	\mathcal{J}_{\text{sun}} = \sum_{k=k_1}^{k_2} \mbf{e}_{\mbf{s}_k}^T \mbf{R}^{-1}_{\mbf{s}_k} \mbf{e}_{\mbf{s}_k},
\end{equation}
and $\mathcal{J}_{\text{reprojection}}$ and $\mathcal{J}_{\text{prior}}$ are defined in \Cref{eq:reprojection_cost,eq:prior_cost}, respectively.
This additional cost term constrains the orientation of the camera, which helps limit drift in the VO result due to orientation error~\cite{Lambert2012-um}.

Since $\mbf{s}_k$ is constrained to be unit length, there are only two underlying degrees of freedom.
We therefore define $\mbf{f}\left(\cdot\right)$ to be a function that transforms a 3D unit vector in camera frame $\cframe{k}$ to a zenith-azimuth parameterization:

\begin{equation} \label{eq:vec-to-az-zen}
	\bbm \theta \\ \phi \ebm
    = \mbf{f} \left( \mbf{s}_k \right)
    = \bbm \text{acos}\left( -s_{k,y} \right) \\ \text{atan2}\left(s_{k,x}, s_{k,z} \right) \ebm
\end{equation}
where $\mbf{s}_k = \bbm s_{k,x} & s_{k,y} &s_{k,z} \ebm^T$.
We can then define the term $\mbf{e}_{\mbf{s}_k} = \mbf{f}\left(\mbf{s}_k\right) - \mbf{f}\left(\mbfhat{s}_k\right)$ to be the error in the predicted sun direction, expressed in azimuth-zenith coordinates, and $\mbf{R}_{\mbf{s}_k}$ to be the covariance of these errors.
While $\mbf{R}_{\mbf{s}_k}$ would generally be treated as an empirically determined static covariance, in our approach we use the per-observation covariance computed using \Cref{eq:bcnn_covar}, which allows us to weight each observation individually according to a measure of its intrinsic quality.
In practice, we also attempt to mitigate the effect of outlier sun predictions by applying a robust Huber loss to the sun measurements in our optimizer.

\section{Experiments}
To train and test Sun-BCNN we used the KITTI odometry benchmark training sequences~\cite{Geiger2013-ky}. Because we rely on the first pose reported by the GPS/INS system, we used the raw (rectified and synchronized) sequences corresponding to each odometry sequence. However, the raw sequence \texttt{2011\_09\_26\_drive\_0067} corresponding to odometry sequence \texttt{03} was not available on the KITTI website at the time of writing, so we omit sequence \texttt{03} from our analysis. In this section, the test datasets simply correspond to each odometry sequence, while the corresponding training datasets consist of the union of the remaining nine sequences.

\subsection{Training Sun-BCNN}

\begin{table*}[]
\centering
\caption{Test Errors for Sun-BCNN on KITTI odometry sequences with estimates computed at every image.}
\label{tab:testCNN}
\begin{threeparttable}
\begin{tabular}{@{}cccccccccccccccc@{}}
         &  & \multicolumn{3}{c}{\textbf{Zenith Error {[}deg{]}}} &  & \multicolumn{3}{c}{\textbf{Azimuth Error {[}deg{]}}} &  & \multicolumn{3}{c}{\textbf{Vector Angle Error {[}deg{]}}} & & \B \\ \cline{3-5} \cline{7-9} \cline{11-13}
\textbf{Sequence} &  & Mean          & Median       & Stdev       &  & Mean         & Median        & Stdev        &  & Mean           & Median          & Stdev & & \textbf{ANEES}\tnote{1}         \T \\ \midrule
\texttt{00}     &  & -2.59  & -1.37  & 5.15           &  & -0.33  & 0.81   & 25.61           &  & 13.56 & 10.31  & 13.14 &  & 1.00 \T \\
\texttt{01}     &  & -12.53 & -8.31  & 10.33          &  & 8.95   & 8.83   & 33.67           &  & 22.16 & 17.85  & 15.00 &  & 1.38 \\
\texttt{02}     &  & -6.13  & -4.26  & 7.38           &  & -1.03  & 0.74   & 37.61           &  & 19.69 & 14.32  & 18.25 &  & 1.40 \\
\texttt{04}     &  & -2.42  & -2.11  & 1.64           &  & -3.89  & -2.18  & 9.14            &  & 5.33  & 3.29   & 6.44  &  & 0.30 \\
\texttt{05}     &  & -4.31  & -2.51  & 6.18           &  & -0.74  & -3.80  & 29.81           &  & 15.66 & 11.33  & 14.80 &  & 1.05 \\
\texttt{06}     &  & -2.48  & -2.52  & 2.27           &  & -12.22 & -17.86 & 25.78           &  & 19.78 & 17.72  & 11.35 &  & 1.93 \\
\texttt{07}     &  & -0.69  & -0.16  & 3.26           &  & 1.25   & 5.98   & 20.27           &  & 12.44 & 10.05  & 9.97  &  & 0.97 \\
\texttt{08}     &  & -4.46  & -1.61  & 8.14           &  & 3.66   & -0.14  & 41.73           &  & 19.90 & 13.30  & 19.59 &  & 1.04 \\
\texttt{09}     &  & -1.35  & -0.75  & 5.60           &  & 4.78   & 2.36   & 23.84           &  & 13.09 & 9.48   & 12.66 &  & 0.73 \\
\texttt{10}     &  & 0.59   & 0.95   & 3.90           &  & 3.64   & 2.61   & 19.15           &  & 11.23 & 8.34   & 9.83  &  & 1.08 \B \\ \midrule
All                &  & -4.01  & -2.26  & 7.06           &  & 0.68   & 0.53   & 32.23           &  & 16.66 & 12.08  & 15.91 & & - &  \\ \bottomrule
\end{tabular}
\begin{tablenotes}
	\item[1] We compute Average Normalized Estimation Error Squared (ANEES) values with all sun directions that fall below a cosine distance threshold of $0.3$ and set $\tau^{-1} = 0.015$.
 \end{tablenotes}
\end{threeparttable}
\end{table*}

\begin{figure*}
    \centering
    \includegraphics[width=0.9\textwidth]{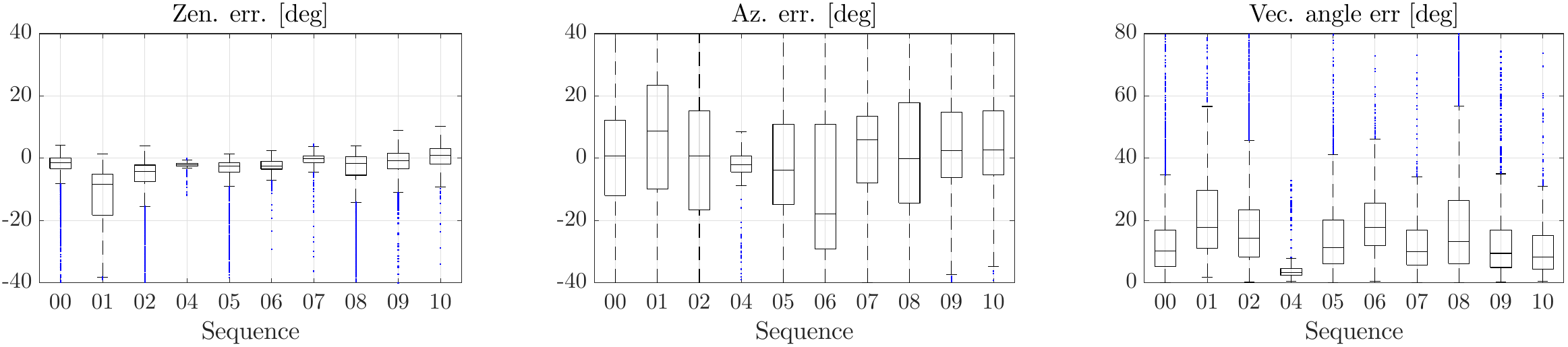}
    \caption{Box-and-whiskers plot of final test errors on all ten KITTI odometry sequences (c.f. \Cref{tab:testCNN}).}
    \label{fig:test_error_whiskers}
\end{figure*}

We implemented our network in Caffe \cite{jia2014caffe} (for the normalization layers, we used the \texttt{L2Norm} layer from the Caffe-SL fork\footnote{\url{https://github.com/wanji/caffe-sl}}) and trained the network using stochastic gradient descent, performing 30,000 iterations with a batch size of 64. This results in approximately 1000 epochs of training on an average of roughly 20,000 images. We set all dropout probabilities to 0.5.

\subsubsection{Data Preparation \& Transfer Learning}
We resized the KITTI images from their original, rectified size of $[1242 \times 378]$ pixels to $[224 \times 224]$ pixels to achieve the image size expected by GoogleLeNet. We experimented with preserving the aspect ratio of the original image (padding zeros to the top and bottom of the resized image), but found that preserving the vertical resolution (as in~\cite{Ma2016-at}) resulted in better test-time accuracy. We performed no additional cropping or rotating of the images.

\subsubsection{Model Precision}
We found an empirically optimal model precision $\tau$ (see \Cref{eq:model_precision}) by optimizing the Average Normalized Estimation Error Squared (ANEES) on test error. In principle, this hyperparameter should be tuned using a validation set, but we omitted this step to keep our training procedure close to that of Ma et al  \cite{Ma2016-at}. We note that the BCNN uncertainty estimates are affected by two significant factors: 1) variational inference is known to underestimate predictive variance \cite{Gal2016UncertaintyThesis}; and 2) we assume the observation noise is homoscedastic. As noted by Gal~\cite{Gal2016UncertaintyThesis}, the BCNN can be made heteroscedastic by learning the model precision during training, but this extension is outside the scope of this work.

\subsection{Testing Sun-BCNN}

Once trained, we analyzed the accuracy and consistency of Sun-BCNN mean ($\mbf{s}_k $) and covariance ($\mbf{R}_{\mbf{s}_k}$) estimates.

\subsubsection{Computing $\mbf{s}_k $}
We evaluated \Cref{eq:sun_direction_mean} (setting $N=25$) and then renormalized the resulting mean vector to preserve unit length.

\subsubsection{Computing $\mbf{R}_{\mbf{s}_k}$}
To obtain the required covariance on azimuth and zenith angles (recall that the BCNN outputs unit-length direction vectors), we sampled the vector outputs, converted them to azimuth and zenith angles using \Cref{eq:vec-to-az-zen}, and then applied \Cref{eq:bcnn_covar}. It is also possible to retain samples in unit vector form, apply \Cref{eq:bcnn_covar}, and then propagate this covariance through a linearized \Cref{eq:vec-to-az-zen}. In this paper we used the former approach, leaving a comparison of these two uncertainty propagation schemes to future work.

\subsubsection{Results}
\label{sec:sun-bcnn-test-results}
\begin{figure}
    \centering
    \begin{subfigure}[b]{0.48\textwidth}
        \includegraphics[width=\textwidth]{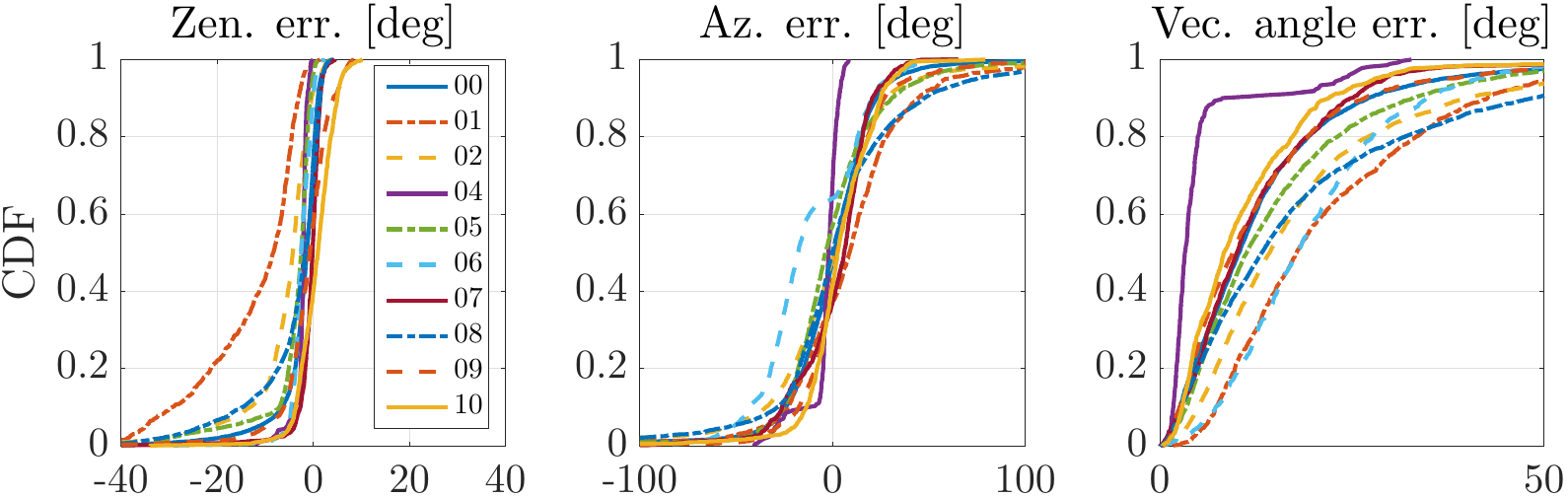}
        \vspace{0.25em}
    \end{subfigure}
    \begin{subfigure}[b]{0.48\textwidth}
        \includegraphics[width=\textwidth]{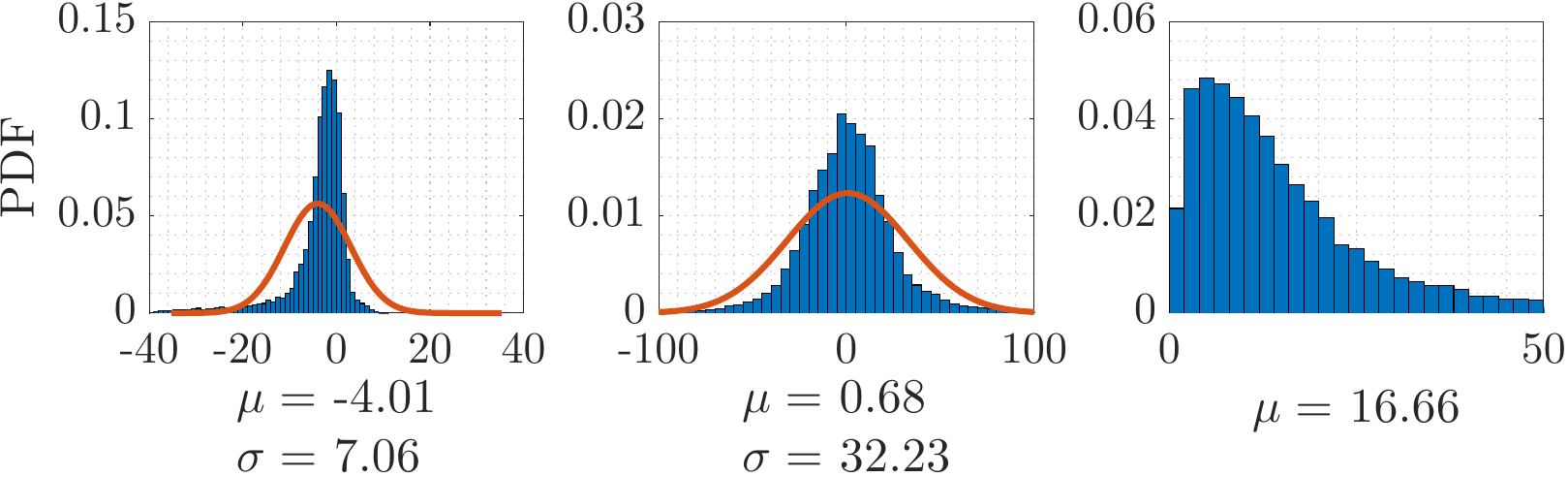}
    \end{subfigure}
    \caption{Distributions of azimuth error, zenith error, and angular distance for Sun-BCNN compared to ground truth over each test sequence. \emph{Top row}: Cumulative distributions of errors for each test sequence individually. \emph{Bottom row:} Histograms and Gaussian fits of aggregated errors.}
    \label{fig:cnn_testerrors}
\end{figure}

\begin{figure}
    \centering
    \includegraphics[width=0.48\textwidth]{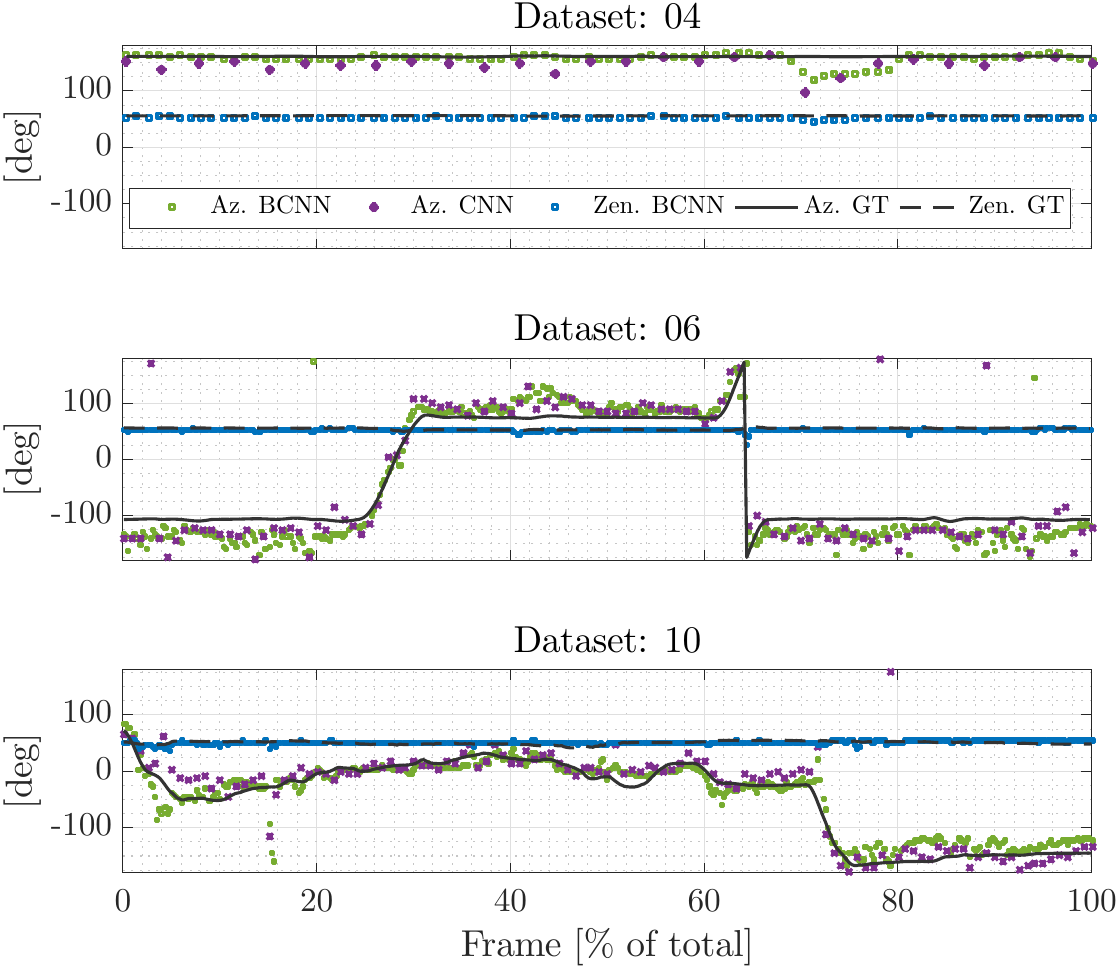}
    \caption{Azimuth (Sun-BCNN and Sun-CNN \cite{Ma2016-at}) and zenith (Sun-BCNN only) predictions over time for KITTI test sequences \texttt{04}, \texttt{06} and \texttt{10}. Sun-CNN is trained and tested on every tenth image, whereas Sun-BCNN is trained and tested on all frames (in our VO experiments, we use the Sun-BCNN predictions of every tenth image to make a fair comparison). }
    \label{fig:error_over_time}
\end{figure}

 \Cref{tab:testCNN} summarizes the test errors numerically, while \Cref{fig:test_error_whiskers,fig:cnn_testerrors} plot the error distributions for azimuth, zenith, and angular distance for all ten KITTI odometry sequences. \Cref{tab:testCNN} also lists the Average Normalized Estimation Error Squared (ANEES) values for each sequence. \Cref{fig:error_over_time} shows three characteristic plots of the azimuth and zenith predictions over time. Sun-BCNN achieved median vector angle errors of less than 15 degrees on every sequence except \texttt{01} and \texttt{06}, which were particularly difficult in places due to challenging lighting conditions. As illustrated in \Cref{fig:cnn_activations}, Sun-BCNN often relies on strong shadows to estimate the sun direction.

\subsection{Visual Odometry with Simulated Sun Sensing}
\label{sec:vo_sim_sun}

In order to gauge the effectiveness of incorporating sun information in each sequence, and to determine the impact of measurement error, we constructed several sets of simulated sun measurements by computing ground truth sun vectors and artificially corrupting them with varying levels of zero-mean Gaussian noise.
We obtained these ground truth sun vectors by transforming the ephemeris vector into each camera frame using ground truth vehicle poses.
We selected our noise levels such that the mean angular error of each simulated dataset was approximately 0, 10, 20, and 30 degrees, and denote each such dataset as ``GT-Sun-0'', ``GT-Sun-10'', ``GT-Sun-20'', and ``GT-Sun-30'', respectively.

\Cref{fig:05-cnn-results-a,fig:05-cnn-results-b,fig:05-cnn-results-c} show the results we obtained using simulated sun measurements on the 2.2 km odometry sequence \texttt{05}, in which the basic VO suffers from substantial orientation drift.
Incorporating absolute orientation information from the simulated sun sensor allows the VO to correct these errors, but the magnitude of the correction decreases as sensor noise increases.
As shown in \Cref{tab:armse}, which summarizes our VO results for all ten sequences, this is typical of sequences where orientation drift is the dominant source of error.

While the VO solutions for sequences such as \texttt{00} do not improve in terms of translational ARMSE, \Cref{tab:armse} shows that rotational ARMSE nevertheless improves on all ten sequences when low-noise simulated sun measurements are included.
This implies that the estimation errors of the basic VO solutions for certain sequences are dominated by non-rotational effects, and that the apparent benefit of the Lalonde method on translational ARMSE in sequence \texttt{00} is likely coincidental.
We speculate that incorporating a motion prior in our VO pipeline may mitigate these additional translational errors, and leave such an investigation to future work.

\begin{table*}[]
\centering
\caption{Comparison of translational and rotational average root mean squared error (ARMSE) on KITTI odometry sequences with and without sun direction estimates at every tenth image. The best result (excluding simulated sun sensing) is highlighted in bold.}
\label{tab:armse}
\begin{threeparttable}
\begin{tabular}{@{}lcccccccccc@{}}
\textbf{Sequence}\tnote{1}     & \texttt{00} & \texttt{01}\tnote{2} & \texttt{02} & \texttt{04} & \texttt{05} & \texttt{06} & \texttt{07} & \texttt{08} & \texttt{09} & \texttt{10} \\ \midrule
\textbf{Length {[}km{]}}       & 3.7   & 2.5    & 5.1   & 0.4  & 2.2   & 1.2   & 0.7   & 3.2   & 1.7   & 0.9   \\ \midrule
\textbf{Trans. ARMSE {[}m{]}}  &       &        &       &      &       &       &       &       &       &       \\
\quad Without Sun & 4.33          & 198.52          & 28.59          & 2.48          & 9.90          & 3.35          & 4.55          & 28.05          & 10.44          & 5.54          \T\B \\
\quad GT-Sun-0    & 5.40          & 114.69          & 23.83          & 2.23          & 4.84          & 3.50          & 1.58          & 31.55          & 8.21           & 3.67          \T \\
\quad GT-Sun-10   & 4.85          & 123.84          & 25.34          & 2.45          & 5.84          & 2.80          & 2.94          & 28.47          & 8.65           & 4.81          \\
\quad GT-Sun-20   & 4.78          & 136.60          & 22.33          & 2.46          & 8.16          & 3.03          & 3.90          & 27.54          & 8.68           & 5.45          \\
\quad GT-Sun-30   & 4.83          & 157.14          & 27.30          & 2.48          & 8.93          & 3.44          & 4.62          & 26.73          & 10.10          & 5.28          \B \\
\quad Lalonde     & \textbf{3.81} & 200.34          & 28.13          & \textbf{2.47} & 9.88          & 3.36          & 4.61          & 29.70          & 10.49          & \textbf{5.48} \T \\
\quad Lalonde-VO  & 4.87          & 199.03          & 29.41          & 2.48          & 9.74          & \textbf{3.30} & 4.52          & 27.82          & 11.06          & 5.59          \B \\
\quad Sun-CNN     & 4.36          & 192.50          & \textbf{26.58} & 2.48          & 8.92          & 3.38          & 4.30          & \textbf{26.99} & 10.15          & 5.58          \T \\
\quad Sun-BCNN    & 4.44          & \textbf{188.46} & 26.89          & 2.48          & \textbf{8.50} & 4.10          & \textbf{4.21} & 27.71          & \textbf{10.13} & 5.61          \\ \midrule
\textbf{Trans. ARMSE (EN-plane) {[}m{]}}                    &       &        &       &      &       &       &       &       &       &       \\
\quad Without Sun & 4.53          & 230.73          & 30.66          & 1.81          & 11.50         & 3.68          & 5.44          & 32.37          & 11.65          & 5.95          \T\B \\
\quad GT-Sun-0    & 3.41          & 136.76          & 24.12          & 1.46          & 3.67          & 3.96          & 1.80          & 21.51          & 7.77           & 3.71          \T \\
\quad GT-Sun-10   & 5.05          & 149.36          & 24.79          & 1.79          & 6.29          & 2.73          & 3.51          & 22.41          & 8.90           & 5.09          \\
\quad GT-Sun-20   & 5.14          & 164.37          & 22.04          & 1.80          & 9.01          & 3.13          & 4.66          & 27.58          & 8.86           & 5.81          \\
\quad GT-Sun-30   & 5.12          & 188.61          & 22.65          & 1.83          & 10.31         & 3.83          & 5.50          & 27.65          & 11.16          & 5.58          \B \\
\quad Lalonde     & \textbf{3.95} & 232.66          & 27.30          & \textbf{1.81} & 11.20         & 3.70          & 5.52          & 27.84          & 11.41          & \textbf{5.87} \T \\
\quad Lalonde-VO  & 5.38          & 231.33          & 33.68          & 1.82          & 11.13         & \textbf{3.61} & 5.42          & 32.24          & 12.41          & 6.00          \B \\
\quad Sun-CNN     & 4.56          & 224.91          & 24.65          & 1.82          & 9.99          & 3.74          & 5.16          & 30.09          & 11.21          & 5.99          \T \\
\quad Sun-BCNN    & 4.68          & \textbf{220.54} & \textbf{23.58} & 1.82          & \textbf{6.70} & 4.78          & \textbf{5.05} & \textbf{26.59} & \textbf{10.97} & 6.03          \\ \midrule
\textbf{Rot. ARMSE $\mbf{(\times 10^{-3})}$ {[}axis-angle{]}} &       &        &       &      &       &       &       &       &       &       \\
\quad Without Sun & 23.88          & 185.30          & 63.18          & 12.97          & 70.18          & 23.24          & 49.96          & 63.13          & 26.77          & 21.54          \T\B \\
\quad GT-Sun-0    & 11.20          & 38.82           & 53.48          & 11.75          & 29.38          & 17.66          & 20.37          & 56.39          & 17.00          & 12.60          \T \\
\quad GT-Sun-10   & 17.05          & 64.51           & 58.78          & 12.86          & 41.47          & 18.90          & 34.05          & 54.89          & 19.71          & 14.26          \\
\quad GT-Sun-20   & 18.84          & 94.65           & 58.03          & 12.91          & 55.39          & 19.67          & 43.34          & 58.82          & 20.99          & 25.87          \\
\quad GT-Sun-30   & 23.40          & 121.21          & 57.79          & 13.01          & 62.73          & 23.96          & 49.92          & 56.74          & 25.63          & 20.15          \B \\
\quad Lalonde     & \textbf{21.10} & 188.06          & 66.02          & \textbf{12.96} & 69.00          & 23.27          & 50.49          & 64.22          & 26.27          & \textbf{20.49} \T \\
\quad Lalonde-VO  & 27.91          & 185.52          & 69.52          & 12.98          & 68.09          & \textbf{22.79} & 49.74          & 65.35          & 28.82          & 22.10          \B \\
\quad Sun-CNN     & 24.05          & 177.45          & \textbf{58.32} & 13.00          & 61.48          & 23.34          & 47.77          & \textbf{60.55} & \textbf{26.19} & 21.99          \T \\
\quad Sun-BCNN    & 26.96          & \textbf{175.21} & 75.02          & 13.00          & \textbf{47.96} & 23.80          & \textbf{47.57} & 62.85          & 26.29          & 20.85          \\ \bottomrule
\end{tabular}
\begin{tablenotes}
	\item[1] Because we rely on the timestamps and first pose reported by the GPS/INS system, we use the raw (rectified and synchronized) sequences corresponding to each odometry sequence. However, the raw sequence \texttt{2011\_09\_26\_drive\_0067} corresponding to odometry sequence \texttt{03} was not available on the KITTI website at the time of writing, so we omit sequence \texttt{03} from our analysis.
    \item[2] Sequence \texttt{01} consists largely of self-similar, corridor-like highway driving which causes difficulties when detecting and matching features using \texttt{libviso2}. The base VO result is of low quality, although we note that including global orientation from the sun nevertheless improves the VO result.
\end{tablenotes}
\end{threeparttable}
\end{table*}


\subsection{Visual Odometry with Vision-based Sun Sensing}
\begin{figure*}
	\centering
	 \begin{subfigure}{0.32\textwidth}
    	\includegraphics[width=\textwidth]{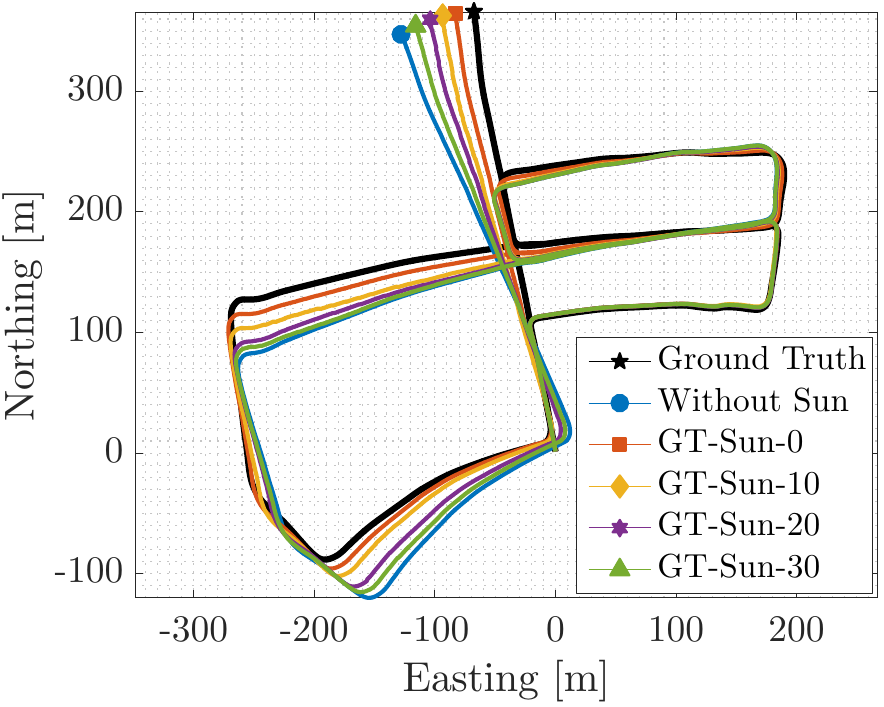}
        \caption{\emph{Simulated sun measurements}: VO trajectories (EN plane).}
        \label{fig:05-cnn-results-a}
    \end{subfigure}
    ~
    \begin{subfigure}{0.31\textwidth}
    	\includegraphics[width=\textwidth]{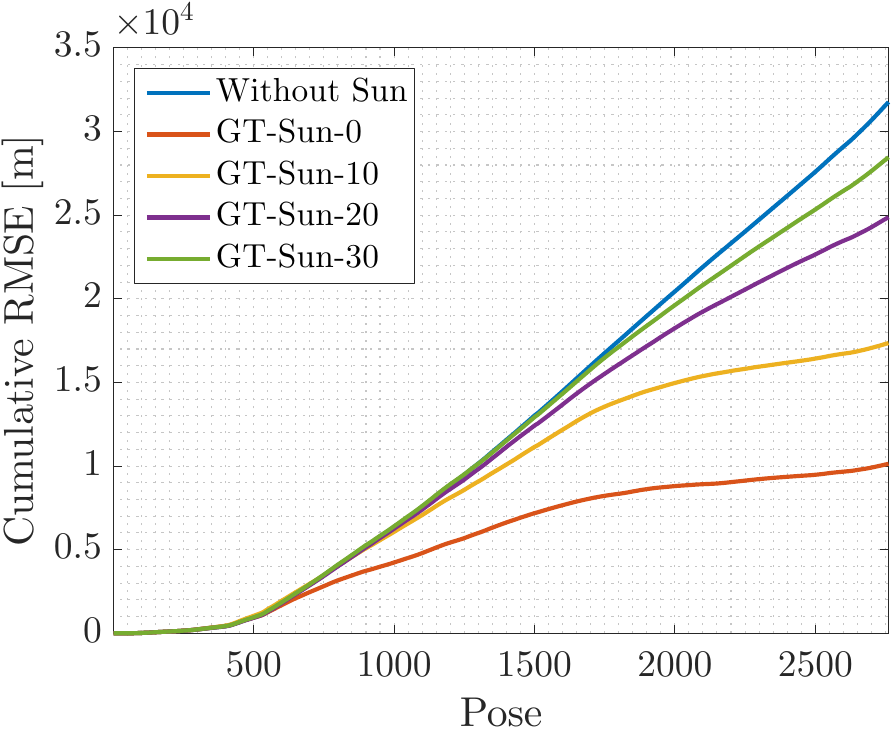}
        \caption{\emph{Simulated sun measurements}: CRMSE of VO translation estimates (EN plane).}
        \label{fig:05-cnn-results-b}
    \end{subfigure}
    ~
    \begin{subfigure}{0.32\textwidth}
    	\includegraphics[width=\textwidth]{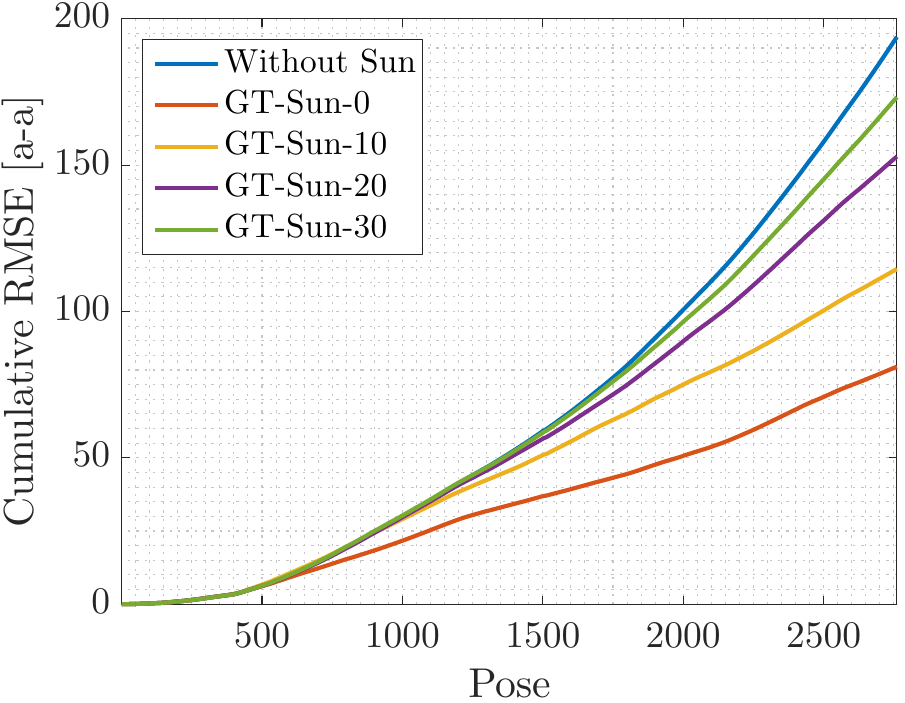}
        \caption{\emph{Simulated sun measurements}: CRMSE of VO rotation estimates.}
        \label{fig:05-cnn-results-c}
    \end{subfigure}
    ~
    \begin{subfigure}{0.32\textwidth}
    	\includegraphics[width=\textwidth]{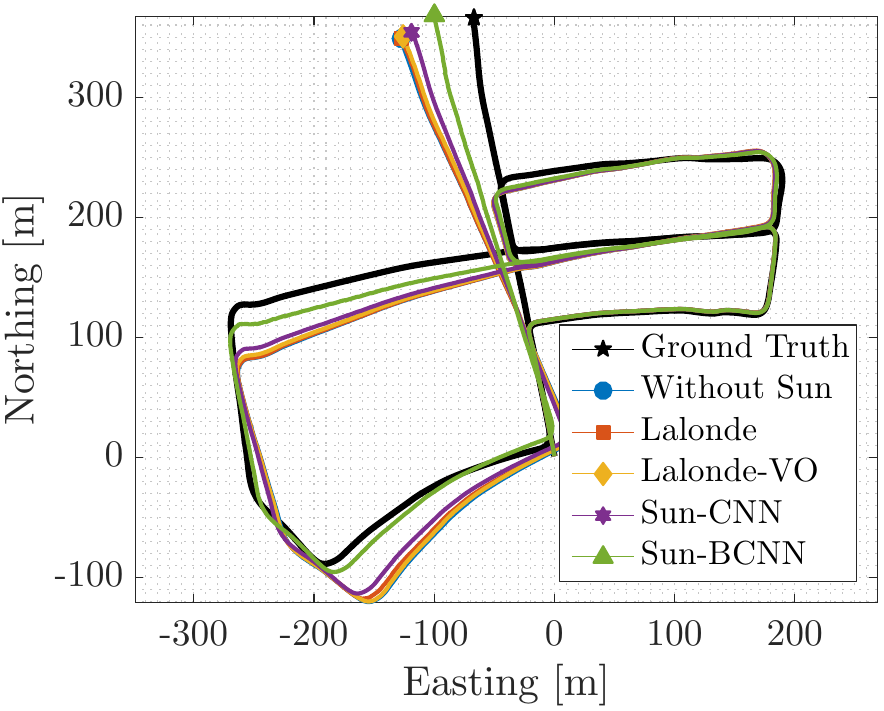}
        \caption{\emph{Estimated sun measurements}: VO trajectories (EN plane).}
        \label{fig:05-cnn-results-d}
    \end{subfigure}
    ~
    \begin{subfigure}{0.31\textwidth}
    	\includegraphics[width=\textwidth]{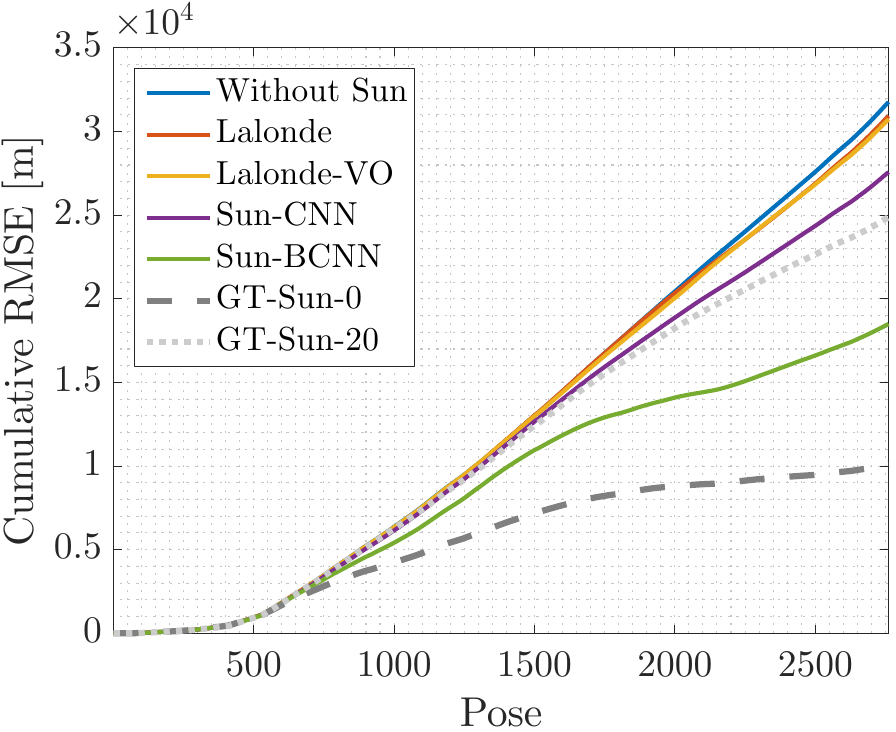}
        \caption{\emph{Estimated sun measurements}: CRMSE of VO translation estimates (EN plane).}
        \label{fig:05-cnn-results-e}
    \end{subfigure}
    ~
    \begin{subfigure}{0.32\textwidth}
    	\includegraphics[width=\textwidth]{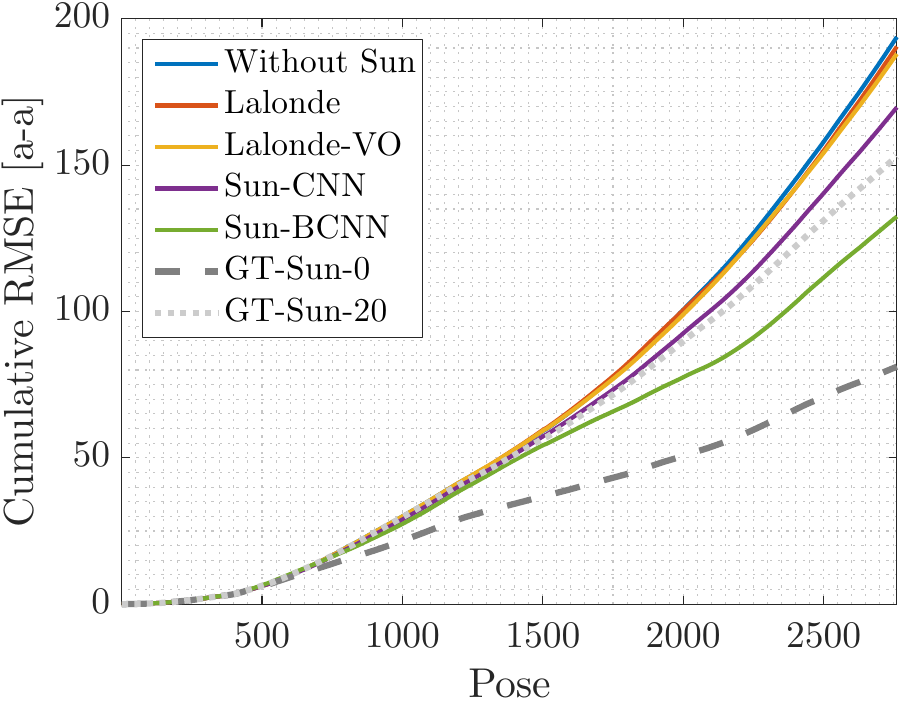}
        \caption{\emph{Estimated sun measurements}: CRMSE of VO rotation estimates.}
        \label{fig:05-cnn-results-f}
    \end{subfigure}
    \caption{VO results for sequence \texttt{05}: top down trajectory plots in the Easting-Northing (EN) plane and Cumulative Root Mean Squared Error (CRMSE) plots for translational and rotational error. \emph{Top row}: Results using a selection of simulated sun measurements of varying accuracy (c.f. \Cref{sec:vo_sim_sun}). \emph{Bottom row}: Results using different sun estimation techniques, with selected simulated measurements added for reference. The sun direction estimates provided by Sun-BCNN significantly improve the VO solution, while the Lalonde~\cite{Lalonde2011-jw}, Lalonde-VO~\cite{Clement2016-ir}, and Sun-CNN~\cite{Ma2016-at} methods provide modest reductions in estimation error. In both simulated and estimated measurements, sun directions are computed at every tenth pose.}
     \label{fig:05-cnn-results}
\end{figure*}

\Cref{fig:05-cnn-results-d,fig:05-cnn-results-e,fig:05-cnn-results-f} show the results we obtained for sequence \texttt{05} using the Sun-CNN of Ma et al.~\cite{Ma2016-at}, which estimates only the azimuth angle of the sun, our Bayesian Sun-BCNN which provides full 3D estimates of the sun direction as well as a measure of the uncertainty associated with each estimate, and the method of Lalonde et al. in its original~\cite{Lalonde2011-jw} and VO-informed~\cite{Clement2016-ir} forms, which provide 3D estimates of the sun direction without reasoning about uncertainty.
A selection of results using simulated sun measurements are also displayed for reference.
All four sun detection methods succeed in reducing the growth of total estimation error on this sequence, with Sun-BCNN reducing both translational and rotational error growth significantly more than the other three methods.
Both Sun-CNN and Sun-BCNN outperform the two Lalonde variants, consistent with the results of Ma et al.~\cite{Ma2016-at} and Clement et al.~\cite{Clement2016-ir}.

\Cref{tab:armse} shows results for all ten sequences using each method.
With few exceptions, the VO results using Sun-BCNN achieve improvements in rotational and translational ARMSE comparable to those achieved using the simulated sun measurements with between 10 and 30 degrees average error.
As previously noted, sequences such as \texttt{00} do not benefit significantly from sun sensing since rotational drift is not the dominant source of estimation error in these cases.
Nevertheless, these results indicate that CNN-based sun sensing is a valuable tool for improving localization accuracy in VO -- an improvement that comes without the need for additional sensors or a specially oriented camera.

\section{Conclusion \& Future Work}
In this work, we have presented Sun-BCNN, a Bayesian CNN applied to the problem of sun direction estimation from a single RGB image in which the sun may not be visible. By leveraging the principled uncertainty estimates of the BCNN, we incorporated the sun direction estimates into a stereo visual odometry pipeline and demonstrated significant reductions in error growth over 21.6 km of urban driving data from the KITTI odometry benchmark. By using a full complement of dropout layers, we were able to train the network using a relatively small training set while achieving a median test error rate of approximately 12 degrees. We stress that although we integrated Sun-BCNN into a visual odometry pipeline in this work, it can just as readily be used to inject global orientation information into any egomotion estimator.

Possible avenues for future work include investigating the effect of cloud cover on sun direction estimates, an analysis of the effect of hyperparameters such as length scale and weight decay on the final model, and the use of multiple cameras with non-overlapping fields of view to compute and combine sun direction estimates from multiple perspectives.

\bibliographystyle{IEEEtran}
\bibliography{sunvo_icra2017.bib}

\end{document}